\documentclass{article} % For LaTeX2e
\usepackage[utf8]{inputenc}
\usepackage{iclr2024_conference,times}

\usepackage{graphicx}
\usepackage{adjustbox}

\usepackage{amsmath}
\usepackage{amsthm}
\usepackage{scalerel}
\NewDocumentCommand\emojijoy{}{\scalerel*{\includegraphics[height=3em]{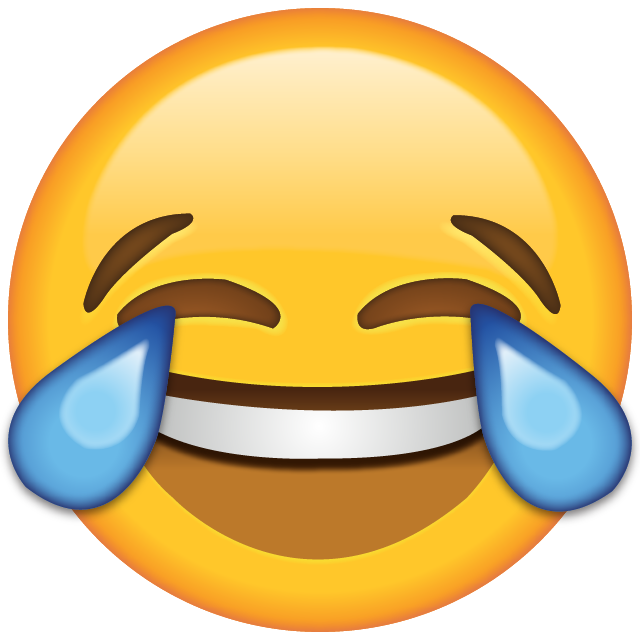}}{joy}}
\NewDocumentCommand\emojimuscle{}{\scalerel*{\includegraphics[height=3em]{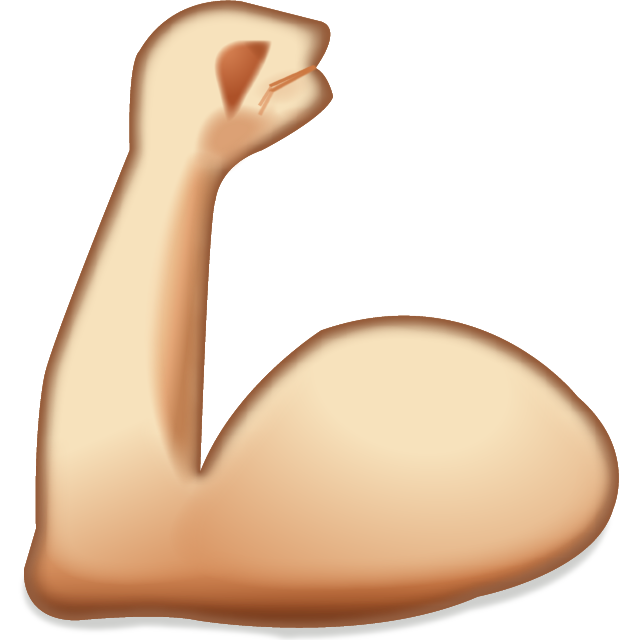}}{muscle}}
\NewDocumentCommand\emojighost{}{\scalerel*{\includegraphics[height=3em]{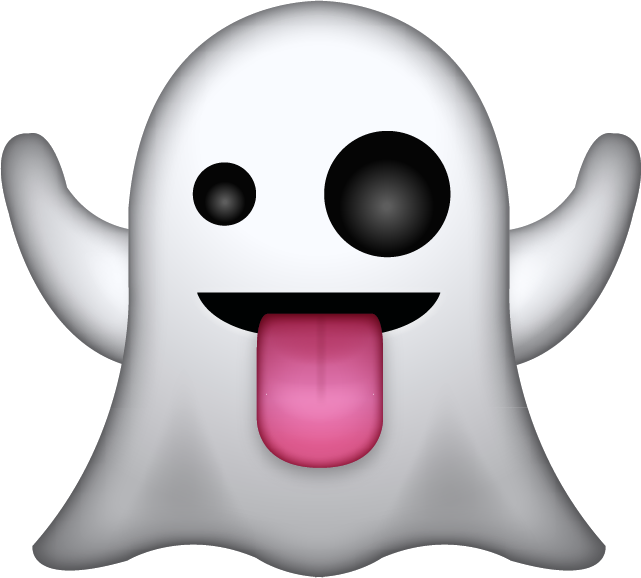}\hspace{1em}}{ghost}}
\NewDocumentCommand\emojicat{}{\scalerel*{\includegraphics[height=3em]{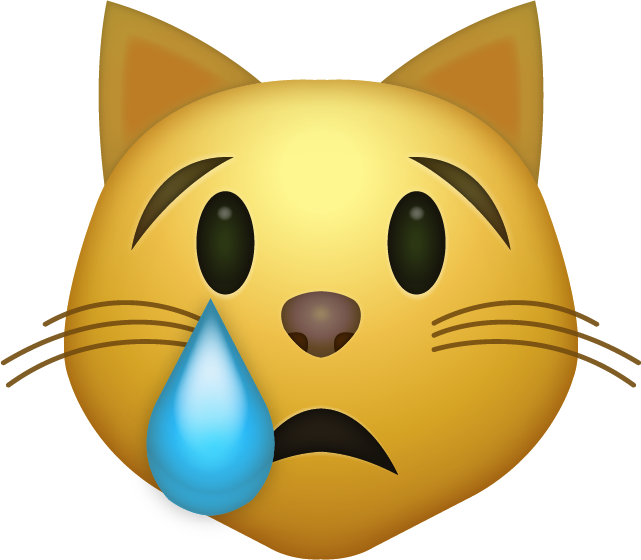}\hspace{0em}}{cat}}
\NewDocumentCommand\emojishrug{}{\scalerel*{\includegraphics[height=3em]{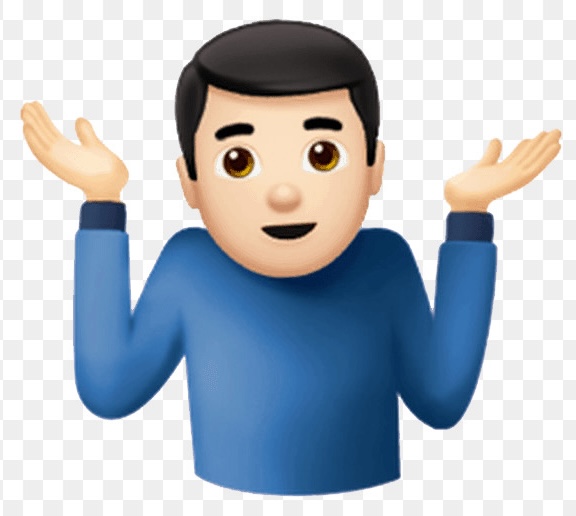}\hspace{0em}}{shrug}}

\usepackage{amsmath,amsfonts,bm}

\def\eqref#1{equation~\ref{#1}}
\def\1{\bm{1}}

\DeclareMathAlphabet{\mathsfit}{\encodingdefault}{\sfdefault}{m}{sl}
\SetMathAlphabet{\mathsfit}{bold}{\encodingdefault}{\sfdefault}{bx}{n}

\usepackage{hyperref}
\usepackage{url}
\usepackage{epigraph}
\usepackage{fdsymbol}

\usepackage{nopageno} % no page numbers for sigbovik

\renewcommand{\paragraph}[1]{\vspace{3pt plus 1pt minus 1pt}\noindent{\bf #1}\;}

\title{Stale Diffusion: Hyper-Realistic 5D Movie Generation using Old-school Methods}

\author{Jo\~{a}o F. Henriques\thanks{Shared first, second, third and last author.},
\textcolor{black!50}{Dylan Campbell}\thanks{Colour blind, so I want a bright red name please. Thank you.},
\textcolor{black!25}{Tengda Han}\thanks{Due to limited contribution, author's name is partially visible.},
\textcolor{black!1}{Samuel}\thanks{For some authors, double-blind review simply doesn't cut it.}\\
International Centre for Sleep Studies\\
Nappington, UK}

\iclrfinalcopy % Uncomment for camera-ready version, but NOT for submission.
\begin{document}

\maketitle

\begin{abstract}
Two years ago, Stable Diffusion achieved super-human performance at generating images with super-human numbers of fingers.
Following the steady decline of its technical novelty, we propose Stale Diffusion, a method that solidifies and ossifies Stable Diffusion in a maximum-entropy state.
Stable Diffusion works analogously to a barn (the Stable) from which an infinite set of horses have escaped (the Diffusion).
As the horses have long left the barn, our proposal may be seen as antiquated and irrelevant.
Nevertheless, we vigorously defend our claim of novelty by identifying as early adopters of the Slow Science Movement, which will produce extremely important pearls of wisdom in the future.
Our speed of contributions can also be seen as a quasi-static implementation of the recent call to pause AI experiments, which we wholeheartedly support.
As a result of a careful archaeological expedition to 18-months-old Git commit histories, we found that naturally-accumulating errors have produced a novel\footnote{Depending on the perspective.} entropy-maximising Stale Diffusion method, that can produce sleep-inducing hyper-realistic 5D video that is as good as one's imagination.
\end{abstract}

%\epigraph{The cake is a lie.}{Anonymous pioneer of CNN technology, Jaffa cake VAT tribunal, 1991}
\epigraph{``Leave the GAN. Take the cannoli.''}{The Godfather, 1972}

\section{Introduction} \label{sec:intro}

Entropy in the observable universe has been steadily increasing since the Big Bang \citep{bigbang}.
Facing the long-term prospects of the heat death of the universe, it is only natural to ask how to reverse entropy \citep{lastquestion,cleanroom}, however this is not exactly an urgent question.

Machine learning recently presented an exciting method to undo the inevitable diffusion (noising) processes in both nature and bit-rotting data.
Two years ago (i.e., about 30 years in both machine learning and dog years), Stable Diffusion emerged as the de facto standard of in silico generation of ad hoc images and video, a veritable magnum opus of modern technology circa 2022.\footnote{I.e., Latin word salad is as relevant for science today as it was 2000 years ago (XXIV A.D.).}

The present authors are only now catching up to this groundbreaking development.
Being enthusiastic adopters of the Slow Science Movement (SSM), today we are still catching up on the printed proceedings of CVPR 2019.\footnote{Director's Cut, with full-length appendices.}
We note this strategy has been adopted by some of the greatest minds of the 20th century, when slacking off at a patent office and experiencing time dilation by extreme boredom sparked a physics revolution \citep{einstein1905elektrodynamik}.

Inspired by this casual attitude towards human development, as well as healthy disposition towards adult napping, we propose Stale Diffusion, a method to generate dream-like 5D video for human consumption while sleeping.
Our Stale Diffusion method starts from a maximum-entropy distribution---the venerable uniform, which promotes fairness across the entire sample space---and implements a reverse diffusion process that, in infinite time, recovers samples from the original data distribution.

We believe that proving the limiting case of infinite iterations brings it in line with the compute requirements of today's SOTA methods.
Our approach is inherently fair, as part of the diffuse--confuse--infuse--reuse--refuse cycle of the textile industry on which the uniform distribution is founded.\footnote{Excluding some non-uniform textiles such as patchwork quilts and grad students' t-shirts.}
By \textit{``stale''}, we are not making a claim of \textit{biscuit-ness} but of \textit{cake-ness} for our model.
That is, our construction hardens when stale, which by the Jaffa Cake Tribunal of 1991 \citep{gov1991jaffa} proves our model to be a cake indeed.
Cakeness is a fundamental property of all important machine learning models \citep{lecun2019deep}.

\section{Unrelated works} \label{sec:related}

\paragraph{I, Robot, by Isaac Asimov.
$\medstar \medstar \medstar \medstar \medstar$
}
Lays out the foundational Three Laws of Robotics. The lesson is that, if implemented perfectly, everything is going to be just fine.

\paragraph{Human Compatible, by Russell.
$\medstar \medstar \medstar \medstar \medwhitestar$
}
Somehow did not read the book above.\footnote{The authors only watched part of the 2004 film with Will Smith, but got the gist of it.}

\paragraph{A 1984 manual for a Russell-Hobbs blender.
$\medstar \medstar \medstar \medwhitestar \medwhitestar$
}
Extremely accessible, translated to 12 languages.
The blender diffused kale and açai very well, but not an~iPhone~3.

\paragraph{Dog by Pablo Picasso.
$\medstar \medstar \medwhitestar \medwhitestar \medwhitestar$
}
It is truly a matter of taste.

\paragraph{Various unrelated works by prominent authors \citep{albanie2017stopping,albanie2018substitute,albanie2019deep,albanie2020stateofartreviewing,albanie2021origin,albanie202223,albanie2023large}. 
$\medstar \medstar \medstar \medstar \medstar$
} By introducing a number of gratuitous self-references, we guarantee
quadratic scaling of citations, thus securing rosy career prospects.

\section{Method} \label{sec:method}

In this section, we've been informed that it's good to have at least one sentence before moving on to the first subsection.
So here it is.

\subsection{Old-school approaches}

To make our method easier to grasp for the younger generations, we present this simple equation of our cost-constrained approach to diffusion maximisation via a Difference of Gaussians operator, which we were assured is as hip now as it was 20 years ago:
\begin{equation} \sup_{y_{o}}\left[\mathrm{DoG}\left(y_{o}\right)\right],\quad \text{s.t. } \,\frac{xz\,\mathrm{bit}}{50\,\mathrm{cent}}\geq1,
\end{equation}
where the ratio is expressed in Ke\$has.
Recently, the community has looked favourably upon splatting when it comes to Gaussians \citep{kerbl3Dgaussians}, but we consider this off-colour when it comes to Difference of Gaussians (DoG) and any other animal-themed operators.

\subsection{Uniforms as maximally-stale distributions}

The uniform includes \emph{all structures} (secret or otherwise) inside it, as it is an envelope for all other distributions.
Thus, we briefly considered titling our paper ``Uniforms are all you need'', but we felt that yet another ``X is all you need'' title is not all that you need.

\subsection{Architecture}

We base our architectural considerations on strict Feng Shui strictures.
As the backbone of our method, instead of a traditional CNN (which has redundant channels and is hardly newsworthy), we use a powerful Transformer.
Our Transformer automatically alternates between its vehicular and anthropomorphic form depending on the needs of the plot of the generated video.\footnote{This behaviour arose spontaneously from pre-training on Saturday morning cartoons.}
Our architecture was implemented following the schema set out in Prof.~Nick Lane's book ``Transformer'' \citep{lane2022transformer} and is therefore fully compatible with the Krebs cycle.
The central equation is given by
\begin{align}
\mathrm{CH_3COSCoA
 + NADH + CO_2} \gets \mathrm{softmax} ( \mathrm{NAD^+} \,\mathrm{HSCoA}^{\mathsf{T}} / \sqrt{d_k}) \, \mathrm{CH_3COCOO^-},
\end{align}
which characterises the decarboxylation of pyruvate via the attention mechanism.

\subsection{Dataset and training}
Our method's training regime is relatively straightforward.
We simply apply the cr-hinge loss to very large collections of TikTok videos, with a crying-joy emoji token appended to each input.
The per-instance loss is therefore given by
\begin{align}
    \ell_\text{cr-hinge} (\hat{y}) &= \max\{0, 1 - y\hat{y}\} \text{, where}\\
    \hat{y} &= f(x \oplus \text{\emojijoy}; \theta),
\end{align}
for a powerful (\emojimuscle) Transformer $f$ with parameters $\theta$, ground-truth labels $y$ direct from TikTok, inputs $x$, and the non-learnable crying-joy parameter \emojijoy.
Experiments with {\scalebox{1.3}\emojicat} were ultimately unsuccessful, and we defer ablations with \emojighost for future work.

\begin{figure}
\centering
\renewcommand{\arraystretch}{1.5}
\setlength{\tabcolsep}{0pt}

\begin{tabular}{p{0.35\textwidth}@{\hskip 6pt}p{0.2\textwidth}p{0.2\textwidth}p{0.2\textwidth}}

Prompt &  & Video output &  \\

\it ``A confident burglar attempts daring heists of archaeological treasures, while being thwarted by the wannabe world police'' & \adjustbox{valign=t}{\includegraphics[width=0.2\textwidth]{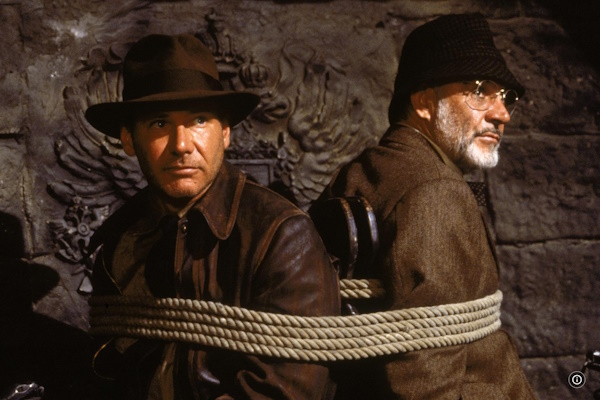}} & \adjustbox{valign=t}{\includegraphics[width=0.2\textwidth]{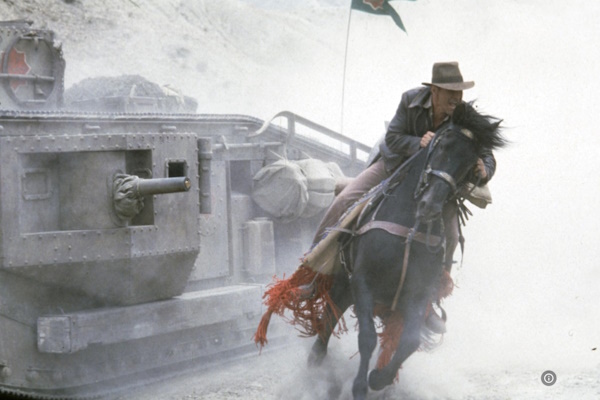}} & \adjustbox{valign=t}{\includegraphics[width=0.2\textwidth]{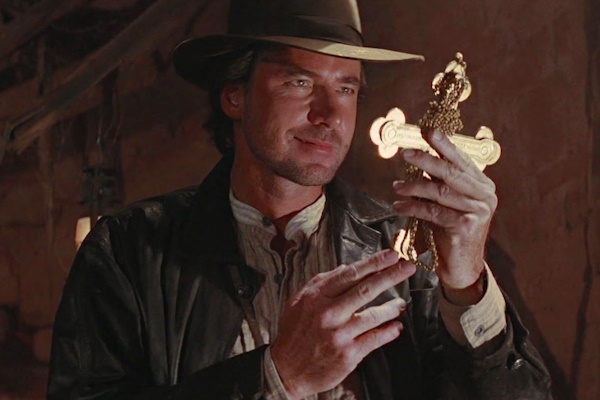}} \\

\it ``A young girl meets sparkly fairies in a magical forest and helps them fight shirtless lumberjacks'' & \adjustbox{valign=t}{\includegraphics[width=0.2\textwidth]{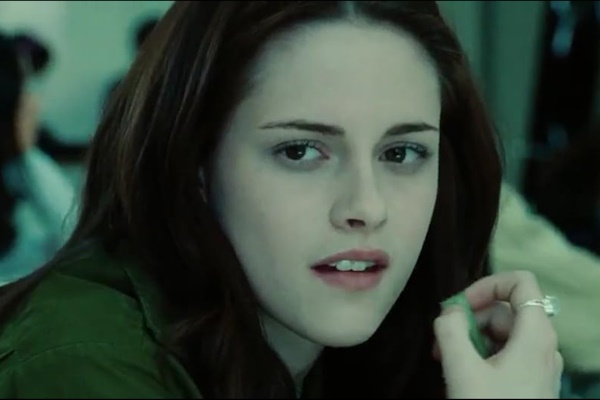}} & \adjustbox{valign=t}{\includegraphics[width=0.2\textwidth]{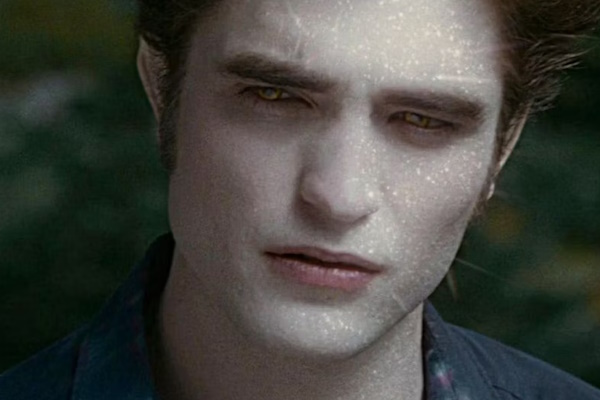}} & \adjustbox{valign=t}{\includegraphics[width=0.2\textwidth]{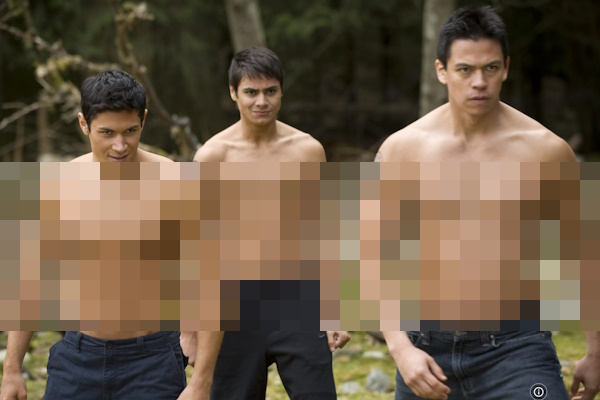}} \\

\it ``Ex Village People band members organise a bombastic reunion party with the help of a young fan'' & \adjustbox{valign=t}{\includegraphics[width=0.2\textwidth]{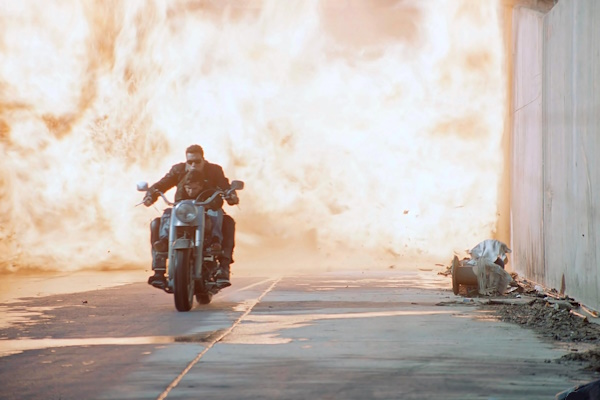}} & \adjustbox{valign=t}{\includegraphics[width=0.2\textwidth]{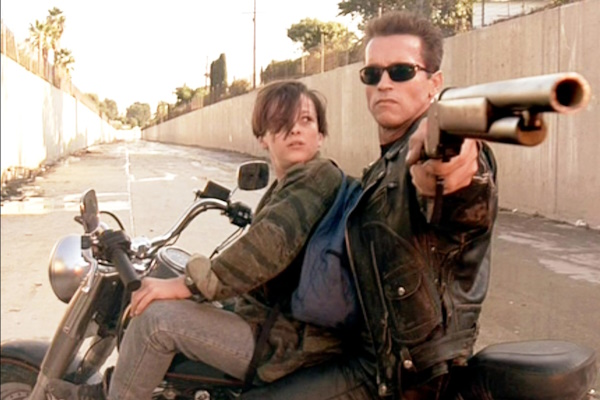}} & \adjustbox{valign=t}{\includegraphics[width=0.2\textwidth]{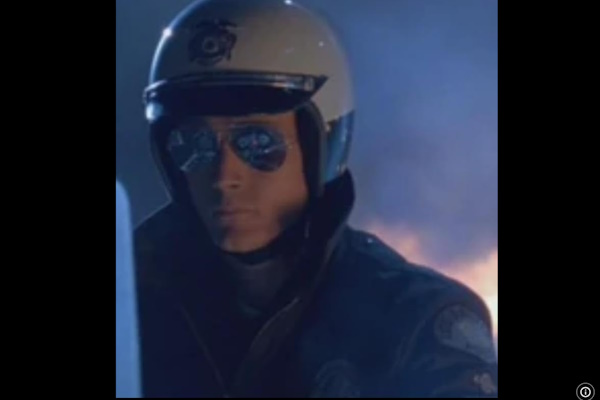}}

\end{tabular}
\vspace{-8pt}
\caption{Example 5D movie-quality videos output by our Stale Diffusion method, downsampled to 2D.
Note that our naturally-aligned network automatically censors types of content that are threatening to insecure researchers, as it is a big prude.}
\label{fig:videos}
\end{figure}

\section{Experiments} \label{sec:experiments}

\paragraph{Implementation details.}
We defer all implementation details to the appendix, which we will release publicly upon our paper's inevitable acceptance in either Nature or Science (we are not picky).
We had some hyper-parameters as well---we believe this is standard practice---but prefer to focus on hyper-realism due to the page limit.

\paragraph{Results.}
Some results from our method are visualised in Fig.~\ref{fig:videos}.
Note that these stills may or may not be identical to photos from IMDb, which we all know is fair game for training generative models.\footnote{If you liked it then you should've put a watermark on it.}
These are, of course, only 2D, due to the limitations of the printed page.\footnote{We tried to use Knuth's \texttt{immersive-vr} \LaTeX{} package, but its ``gaze and pinch'' interface led to some embarrassing pinching incidents.}
Like all great machine learning experiments, \emph{you had to be there to experience it}.

\section{Conclusion} \label{sec:conclusion}
We presented Stale Diffusion, a natural limiting case of Stable Diffusion that addresses its main shortcoming: that it was not proposed by ourselves.
One limitation of our work is that it only applies to the 5 standard human senses, while leaving out the other one that allows you to see Bruce Willis.
There is at least one other significant limitation, but we leave that as a challenge for the reviewers to unearth.
In future work we would like to extend it to more than 5 dimensions, and experiment with mess-up regularisation, an arctangent learning rate schedule, and train--test contamination.

\paragraph{Acknowledgements.} The authors would like to thank ChatGPT for its insightful comments and heartfelt words of encouragement.

\bibliography{sigbovik24}
\bibliographystyle{iclr2024_conference}

\section*{Author biographies}

\textbf{J. F. Henriques}
has published numerous opinion pieces in comment boxes of sensationalist newspapers, which have been cited by a large number of angry responses.
He is the recipient of ``mom's favourite child'' award (joint winner), and currently teaches in the school of life.

\textbf{D. Campbell}
would rather use this space to argue for colour blind review.
The biographee has long been a proponent of greyscale peer review, but has thus far been unsuccessful at having such a motion passed (or indeed considered) at a PAMI-TC meeting.

\textbf{T. Han}
Sure, here is a short bio for the researcher T. Han. From his humble beginnings in a small town, T. Han delved into the extraordinary mysteries of space and time, but a bit far from the realms that Einstein had charted. Do you need help with anything else?

\textbf{\textcolor{black!1}{Samuel}}
\emojighost {\scalebox{1.1}{\emojishrug}

\appendix
\section{Appendix}
\section{Proof of Theorem 1}\label{sec:thm1}
\begin{proof}
Our proof of Theorem 1 works by contradiction. We give a wrong proof of theorem 1 in section~\ref{sec:thm1-wrong}. By first-order logic, since the proof is wrong, then the theorem must be true.
\end{proof}

\section{Proof that Theorem 1 is wrong}\label{sec:thm1-wrong}
\begin{proof}
We start with the main assumptions of Theorem 1. As the main assumptions are mutually incompatible, we find that Theorem 1 cannot be true.
\end{proof}

\section{Proof that the Proof that Theorem 1 is wrong is wrong}\label{sec:thm1-wrong}
\begin{proof}
We start with the main assumptions of the proof that Theorem 1 is wrong.
We then attempted to apply De Morgan's laws.
However, we get distracted by his interesting Wikipedia page and our proof falters.
Did you know that he played the flute recreationally?
\end{proof}

\section{Implementation details}
We leave all implementation details to our reference implementation, which will be uploaded to GitHub upon acceptance of our paper to a major scientific journal (\emph{or else}).

\end{document}